\documentclass[11pt,a4paper]{article}
\usepackage[hyperref]{acl2018}
\usepackage{times}
\usepackage{latexsym}
\usepackage{graphicx}
\usepackage{url}

\aclfinalcopy 

\title{An End-to-end Approach for Handling Unknown Slot Values \\ in Dialogue State Tracking}

\author{Puyang Xu \qquad \qquad \qquad \qquad \qquad \qquad Qi Hu\textsuperscript{\textdagger} \\
\centerline{Mobvoi AI Lab, Redmond, WA} \\
 \textsuperscript{\textdagger}University of Washington, Seattle, WA\\
  {\tt \{puyangxu, qihuchn\}@gmail.com} \\}

\date{}
\begin{document}
\maketitle
\begin{abstract}
We highlight a practical yet rarely discussed problem in dialogue state tracking (DST), namely handling unknown slot values. Previous approaches generally assume predefined candidate lists and thus are not designed to output unknown values, especially when the spoken language understanding (SLU) module is absent as in many end-to-end (E2E) systems. We describe in this paper an E2E architecture based on the pointer network (PtrNet) that can effectively extract unknown slot values while still obtains state-of-the-art accuracy on the standard DSTC2 benchmark. We also provide extensive empirical evidence to show that tracking unknown values can be challenging and our approach can bring significant improvement with the help of an effective feature dropout technique. 
\end{abstract}
\footnotetext{The first author is now with Facebook. Qi contributed to the work during an internship at Mobvoi.}
\section{Introduction}
\label{intr}
A dialogue state tracker is a core component in most of today's spoken dialogue systems (SDS). The goal of dialogue state tracking (DST) is to monitor the user's intentional states during the course of the conversation, and provide a compact representation, often called the dialogue states, for the dialogue manager (DM) to decide the next action to take. 

In task-oriented dialogues, or {\it slot-filling} dialogues in the simplistic form, the dialogue agent is tasked with helping the user achieve simple goals such as finding a restaurant or booking a train ticket.  As the name itself suggests, a slot-filling dialogue is composed of a predefined set of slots that need to be filled through the conversation. The dialogue states in this case are therefore the values of these slot variables, which are essentially the search constraints the DM has to maintain in order to perform the database lookup.

Traditionally in the research community, as exemplified in the dialogue state tracking challenge (DSTC)~\cite{Williams:13}, which has become a standard evaluation framework for DST research, the dialogues are usually constrained by a fixed {\it domain ontology}, which essentially describes in detail all the possible values that each predefined slot can take. Having access to such an ontology can simplify the tracking problem in many ways, however, in many of the SDS applications we have built in the industry, such an ontology was not obtainable. Oftentimes, the back-end databases are only exposed through an external API, which is owned and maintained by our partners. It is usually not possible to gain access to their data or enumerate all possible slot values in their knowledge base. Even if such lists or dictionaries exist, they can be very large in size and highly dynamic (e.g. new songs added, new restaurants opened etc.). It is therefore not amiable to many of the previously introduced DST approaches, which generally rely on classification over a fixed ontology or scoring each slot value pairs separately by enumerating the candidate list.

In this paper, we will therefore focus on this particular aspect of the DST problem which has rarely been discussed in the community -- namely how to perform state tracking in the absence of a comprehensive domain ontology and how to handle unknown slot values effectively.

It is worth noting that end-to-end (E2E) modeling for task-oriented dialogue systems has become a popular trend \citep{Williams:16, Zhao:16, Li:17, Liu:17, Wen:17}, although most of them focus on E2E policy learning and language generation, and still rely on explicit dialogue states in their models. While fully E2E approaches which completely obviate explicit DST have been attempted \citep{Bordes:16, Eric:17a, Eric:17b, Dhingra:17}, their generality and scalability in real world applications remains to be seen. In reality, a dedicated DST component remains a central piece to most dialogue systems, even in most of the proclaimed E2E models.

E2E approaches for DST, i.e. joint modeling of SLU and DST has also been presented in the literature \citep{Henderson:14b, Henderson:14c, Mrksic:15, Zilka:15, Perez:17, Mrksic:17}. In these methods, the conventional practice of having a separate spoken language understanding (SLU) module is replaced by various E2E architectures that couple SLU and DST altogether. They are sometimes called word based state tracking as the dialogue states are derived directly from word sequences as opposed to SLU outputs. In the absence of SLU to generate value candidates, most E2E trackers today can only operate with fixed value sets. To address this limitation, we introduce an E2E tracker that allows us to effectively handle unknown value sets. The proposed solution is based on the recently introduced pointer network (PtrNet) \citep{Vinyals:15}, which essentially performs state tracking in an extractive fashion similar to the sequence labeling techniques commonly utilized for slot tagging in SLU~\citep{Tur:11}.

Our proposed technique is similar in spirit as the recent work in \citep{Rastogi:18}, which also targets the problem of unbounded and dynamic value sets. They introduce a sophisticated candidate generation strategy followed by a neural network based scoring mechanism for each candidate. Despite the similarity in the motivation, their system relies on SLU to generate value candidates, resulting in an extra module to maintain and potential error propagation as commonly faced by pipelined systems.

The contributions of this paper are three-folds: Firstly, we target a very practical yet rarely investigated problem in DST, namely handling unknown slot values in the absence of a predefined ontology. Secondly, we describe a novel E2E architecture without SLU based on the PtrNet to perform state tracking. Thirdly, we also introduce an effective {\it dropout} technique for training the proposed model which drastically improves the recall rate of unknown slot values.

The rest of the paper is structured as follows: We give a brief review of related work in the field in Section~\ref{sec:dst_review} and point out its limitations. The PtrNet and its proposed application in DST are described in Section~\ref{sec:ptr_net}. In Section~\ref{sec:cls_module}, we demonstrate some caveats regarding the use of PtrNet and propose an additional classification module as a complementary component. The targeted dropout technique, which can be essential for generalization on some datasets, are described in Section~\ref{sec:dropout}. Experimental setup and results are presented in Section~\ref{sec:expr}, followed by conclusions in Section~\ref{sec:conclusion}.

\section{Dialogue State Tracking}
\label{sec:dst_review}
In DSTC tasks, the dialogue states are defined as a set of search constraints (i.e. informable slots or goals) the user specified through the dialogue and a set of attribute questions regarding the search results (i.e. requestable slots or requests). The DST component is expected to track the values of the aforementioned slots taking into account the current user utterance as well as the entire dialogue context. As mentioned in the previous section, the values each slot variable can take is specified beforehand through an ontology. This is a hidden assumption that previous techniques usually rely upon implicitly and also what motivates our work in this paper. \\

\noindent\textbf{Discriminative DST}
While generative models aiming at modeling the joint distribution of dialogue states and miscellaneous evidences have been a popular modeling choice for DST for many years, the scalability issue resulting from large state spaces has limited the broader application of this family of models, despite the success of various approximation techniques.

The discriminative methods, on the other hand, directly model the posterior distribution of dialogue states given the evidences accumulated through the conversation history. Models such as maximum entropy \citep{Metallinou:13} and particularly the more recent deep learning based models \citep{Henderson:14b, Henderson:14c, Zilka:15, Mrksic:15, Mrksic:17, Perez:17} have demonstrated state-of-the-art results on public benchmarks. Such techniques often involve a multi-class classification step at the end (e.g. in the form of a softmax layer) which for each slot predicts the corresponding value based on the dialogue history. Sometimes the multi-class classification is replaced by a binary prediction that decides whether a particular slot value pair was expressed by the user, and the list of candidates comes from either a fixed ontology or the SLU output. \\

\noindent\textbf{E2E DST} Previous work has also investigated joint modeling strategies merging SLU and DST altogether. In this line of work, the SLU module is removed from the standard SDS architecture, resulting in reduced development cost and alleviating the error propagation problem commonly affecting cascaded systems. 

In the absence of SLU providing fine-grained semantic features, the E2E approaches these days typically rely on variants of neural networks such as recurrent neural networks (RNN) or memory networks~\citep{Weston:14} to automatically learn features from the raw dialogue history. The deep learning based techniques cited in the previous subsection generally fall into this category.  \\

\noindent\textbf{Current Limitations} In short, most of the previous DST approaches, particularly E2E ones, are not designed to handle slot values that are not known to the tracker.
 
As we have described in the introduction, the assumption that a predefined ontology exists for the dialogue and one can enumerate all possible values for each slot is often not valid in real world scenarios. Such an assumption has implicitly influenced many design choices of previous systems. The methods based on classification or scoring each slot value pair separately can be very difficult to apply when the set of slot values is not enumerable, either due to its size or its constantly changing nature, especially in E2E models where there is no SLU module to generate an enumerable candidate list for the tracker.

It is important to point out the difference between {\it unseen} states and {\it unknown} states, as previous work has tried to address the problem of unseen slot values, i.e. values that were not observed during training. E2E approaches in particular, frequently employ a featurization strategy called {\it delexicalization}, which replaces slots and values mentioned in the dialogue text with generic labels. Such a conversion allows the models to generalize much better to new values that are infrequent or unseen in the training data. However, such slot values are still expected to be known to the tracker, either through a predefined value set or provided by SLU, otherwise the delexicalization cannot be performed, nor can the classifier properly output such values.

\section{Pointer Network}
\label{sec:ptr_net}

In this section, we briefly introduce the PtrNet~\citep{Vinyals:15}, which is the main basis of the proposed technique, and how the DST problem can be reformulated to take advantage of the flexibility enabled by such a model.

In the PtrNet architecture, similar as other sequence-to-sequence (seq2seq) models, there is an encoder which takes the input and iteratively produces a sequence of hidden states corresponding to the feature vector at each input position. There is also a decoder which generates outputs with the help of the weighted encoded states where the weights are computed through attention. Here, instead of using softmax to predict the distribution over a set of predefined candidates, the decoder directly normalizes the attention score at each position and obtains an output distribution over the input sequence. The index of the maximum probability is the pointed position, and the corresponding element is selected as decoder output, which is then fed into next decoding step. Both the encoder and decoder are based on various RNN models, capable of dealing with sequences of variable length.


The PtrNet specifically targets the problems where the output corresponds to positions in the input sequence, and it is widely used for seq2seq tasks where some kind of {\it copying} from the input is needed. Among its various applications, machine comprehension (a form of question answering), such as in~\citep{Wang:16}, is the closest to how we apply the model to DST.

The output of DST, same as in machine comprehension, is a word segment in the input sequence most of the time, thus can be naturally formulated as a {\it pointing} problem. Instead of generating longer output sequences, the decoder only has to predict the starting index and the ending index in order to identify the word segment.

More specifically, words are mapped to embeddings and the dialogue history ${w_0, w_1, ..., w_t}$ up to the current turn $t$ is bidirectionally encoded using LSTM models. To differentiate words spoken by the user versus by the system, the word embeddings are further augmented with speaker role information. Other features, such as the entity type of each word, can also be fed into the encoder simultaneously in order to extract richer information from the dialogue context.

The encoded state at each position can then be denoted as $h_i$, which is the concatenation of forward state and backward state ($[h^f_i, h^b_i]$). The final forward state $h^f_t$ is used as the initial hidden state of the decoder. We use a special symbol denoting the type of slot (e.g. $<$food$>$) as the first decoder input, which is also mapped to a trainable embedding $E_{type}$. Therefore, the starting index $s^0$ of the slot value is computed as the following, where $u_i^0$ is the attention score of the $i_{th}$ word in the input against the decoder state $d^0$.
$$d^0 = LSTM(h^f_t, E_{type})$$
$$u_i^0 = v^T\tanh (W_hh_i + W_dd_0)$$
$$a_i^0 = \exp(u_i^0) / \sum_{j=0}^{t} \exp(u_j^0)$$
$$s^0 = \arg\max_i a_i^0$$
The attention scores at the second decoding step are computed similarly as below, where $E_{w_{s^0}}$ is the embedding of the word at the selected starting position, and the ending position $s^1$ can be obtained in the same way as $s^0$.
$$d^1 = LSTM(d^0, E_{w_{s^0}})$$
$$u_i^1 = v^T\tanh (W_hh_i + W_dd^1)$$

Note that there is no guarantee that $s^1 > s^0$, although most of the time the model is able to identify consistent patterns in the data and therefore output reasonable word segments. When $s^1 < s^0$, it is often a good indication that the answer does not exist in the input (such as the {\it none} slot in DSTC2).\footnote{It is the backoff strategy we take in our experiments on DSTC2.} Depending on the nature of the task, it is certainly possible to set a constraint at the second decoding step, forcing $s^1$ to be larger than $s^0$.

One can clearly see how the described model can handle unknown slot values -- as long as they are mentioned explicitly during the dialogue, we have a chance of finding them. Compared with previous approaches, which all require some kind of candidate lists, the proposed technique takes a different perspective on DST: For most slots in dialogue systems, tracking up-to-date values in a dialogue is not very different from tagging slots in a user query. While sequence labeling models such as conditional random field (CRF) has proven to be a great fit for slot tagging, the same formulation may as well be used for DST.

\begin{figure}
\centering
        \includegraphics[totalheight=6cm]{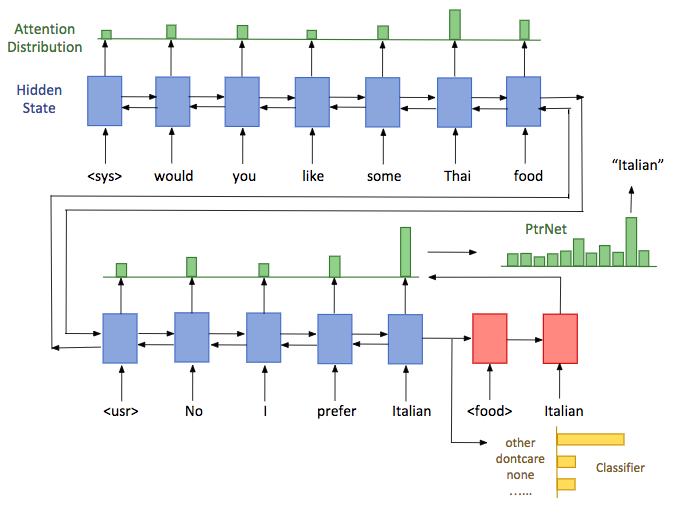}
    \caption{An illustration of the proposed PtrNet based architecture for DST. The classifier outputs ``other" indicating the decision should be made by PtrNet; The decoder (red) in PtrNet is predicting the ending word of the slot value given the predicted starting word via attention against the encoded states (blue).}
    \label{fig:combined}
\end{figure}

\section{Rephrasing and Non-pointable Values}
\label{sec:cls_module}
Our PtrNet based architecture works by directly pinpointing in the conversation history the slot value that the user expressed in its surface form. The model is totally unaware of the different ways of referring to the same entity. Therefore, the derived dialogue states may not have canonical forms that are consistent with the values in the backend database, making it more difficult to retrieve the correct results. A good example from the DSTC2 dataset is the price slot which can take the reference value "moderate", in the actual dialogues however, they are frequently expressed as "moderately priced", causing problems for searching the database and also computing accuracy.

While such a problem can be easily remedied by an extra {\it canonicalization} step (setting dialogue states to standard forms) before performing the database lookup, it is a much bigger problem if the slot value is not indicated explicitly by any particular word or phrase in the dialogue history, we describe these slot  values as {\it non-pointable}. To give an example, in DSTC tasks, the special {\it none} value is given when the user has not specified any constraint for the slot. While this information can be easily inferred from the dialogue, it is not possible to point to any specific word segment in the sentence as the corresponding slot value. The same problem also exists for the {\it dontcare} value in DSTC, which implies that the user can accept any values for a slot constraint.

To address this issue, we add a classification component into our neural network architecture to handle non-pointable values. For each turn of the dialogue, the classifier makes a multi-class decision on whether the target slot should take any of the non-pointable values (e.g. dontcare or none) or it should be processed by the PtrNet. 

As illustrated in Figure~\ref{fig:combined}, the final forward state out of the dialogue encoder is used as the feature vector for the classification layer, which is trained with cross entropy loss and jointly with the PtrNet. 

The best choice of the set of values to be handled by the classifier may not be obvious. In most cases both the classifier and the PtrNet are capable of extracting the correct slot value, although they both offer unique advantages over the other. Table~\ref{tab:cls_vs_ptr} briefly summarizes the pros and cons of each model. 

\begin{table}[t!]
\begin{center}
\begin{tabular}{|c|c|c|}
\hline & \bf Classifer & \bf PtrNet \\ \hline
Rephrasing & Yes & *Yes \\
none, dontcare, etc & Yes & No \\
ASR errors & Hard & Hard \\
Unknown values & No & Yes \\
\hline
\end{tabular}
\end{center}
\caption{Classifier vs. Pointer network in handling various difficult conditions. *PtrNet requires post-normalization to handle rephrasing. }
\label{tab:cls_vs_ptr}
\end{table}

The proposed combined architecture, taking the best of both worlds, is similar to the pointer-generator model introduced in ~\citep{See:17} for abstractive text summarization. In their approach the PtrNet is also augmented with a classification based word generator, and the model can choose  to generate words from a predefined vocabulary or copy words from the input. Other {\it classify-and-copy} mechanisms have also been explored in ~\citep{Gu:16, Gulcehre:16, Eric:17a}, and demonstrated improved performance on various seq2seq tasks such as summarization and E2E dialogue generation. \footnote{The copy-augmented model in~\citep{Eric:17a} also outputs API call parameters (which are essentially dialogue states) in a seq2seq fashion, including unknown parameters by copying from dialogue history, although the work focuses entirely on dialogue generation.} As we have shown in this paper, DST can also be formulated to incorporate such copying mechanisms, allowing itself to handle unknown slot values as well.

\section{Targeted Feature Dropout}
\label{sec:dropout}
Feature dropout is an effective technique to prevent feature co-adaption and improve model generalization~\citep{Hinton:12}. It is most widely used for neural network based models but may as well be utilized for other feature based models. Targeted feature dropout however, was introduced in~\citep{Xu:14} to address a very specific co-adaptation problem in slot filling, namely insufficient training of word context features. 

For slot filling, this problem often occurs when 1) the dictionary (a precompiled list of possible slot values) covers the majority of the slot values in the training data, or 2) most slot values repeat frequently resulting in insufficient tail representations. In both cases, the contextual features tend to get severely under-trained and as a result the model is not able to generalize to unknown slot values that are neither in the dictionary nor observed in training.

\begin{figure}
\centering
        \includegraphics[totalheight=5cm]{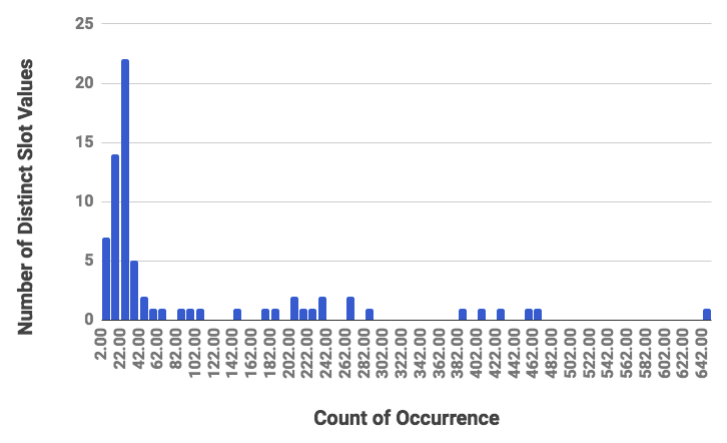}
    \caption{Histogram of {\it food type} slot on DSTC2 training data.}
    \label{fig:hist}
\end{figure}

The way our architecture works essentially extracts slot values in the same way as in slot filling, although the goal is to identify slots considering the entire dialogue context rather than a (usually) single user query. The same problem can also happen for DST if training data are not examined carefully. As an example, the DSTC2 task comes with a fixed ontology, it is not originally designed to track unknown slots (see the OOV rate in Table~\ref{tab:oov test}). Taking a closer look at the data, as shown in the histogram in Figure~\ref{fig:hist}, the majority of the food type slot appears more than 10 times in the training data. As a result, the model oftentimes only learns to memorize these frequent slot values, and not the contextual patterns which can be more crucial for extracting slot values not known in advance.

To alleviate the generalization issue, we adapt the targeted dropout trick to work with our neural network based architecture. Instead of randomly disabling unigram and dictionary features for CRF models as done in the original work, we randomly set to zero the input word embeddings that correspond to the slot values in the dialogue utterances. For example, the {\it italian} food type in DSTC2 appears almost 500 times in the training data. During training, every time ``italian" gets mentioned in the dialogue as the labeled user goal, we turn off the word embedding of ``italian" in the model input with some probability, forcing the model to learn from the context to identify the slot value. Dictionary features are not used in our experiments, otherwise they can be turned off similarly.

As we will show later in the results, this proves to be a particularly effective yet simple trick for improving generalization to unknown slot values, without sacrificing accuracy for the known and observed ones.

\section{Experiments and Results}
\label{sec:expr}
\subsection{Datasets}
We conduct our experiments on the DSTC2 dataset~\citep{Henderson:14a}, and on the bAbI dialogue dataset as used in~\citep{Bordes:16}.

The DSTC2 dataset is the standard DST benchmark comprised of real dialogues between human and dialogue systems. We are mainly interested in tracking user goals, whereas the other two components of the dialogue state, namely search methods and requested slots, are not concerned with unknown slot values, and thus are not the focus in this paper. Meanwhile, the non-pointable values, none and dontcare, constitute a significant portion in DSTC2. Overall almost 60\% of the user goals are labeled as either none or dontcare, the two predominant non-pointable values, it is therefore particularly suitable for evaluating our proposed hybrid architecture. 

An important part of our experimental evaluation is to demonstrate our ability to identify unknown slots. Although it happens frequently in real world situations, the original DSTC2 dataset does not suffer from this particular problem -- on the test data, there are no unknown values that we have not observed in training for all of the three slot types. To conduct our investigation, we pick the food type slot to simulate unknown values. Specifically, we randomly select about 35\% of the food types in the training set (26 out of 74) as unknown and discard all the training instances where the correct food type is one of the 26 unknown types that we selected. The statistics of the resulting dataset is shown in Table~\ref{tab:oov test}.

\begin{table}[t!]
\begin{center}
\begin{tabular}{|c|cl|}
\hline & \bf Original & \bf New \\ \hline
\#food types in train    & 74 & 48 \\
\#train instances & 11677 & 8546 \\
\#test instances & 9890 & 9890 \\
OOV food types in test (\%) & 0 & 30.4 \\
\hline
\end{tabular}
\end{center}
\caption{\label{font-table} Statistics of the new modified DSTC2 dataset with unknown food types.  About 27\% of the training instances are discarded. The test set remains the same. }
\label{tab:oov test}
\end{table}

On the other hand, the bAbI dialogue dataset is initially designed for evaluating E2E goal oriented dialogue systems and has not been used specifically for DST. The model is expected to predict both the system utterances and the API calls to access the database. We notice that the parameters of the API calls are essentially the dialogue states at the point of the dialogue, it may as well be used as a dataset for measuring the accuracy of the state tracker. We therefore convert Task 5 of the bAbI dataset, which is the full dialogue combination of Tasks 1-4, into a DST dataset for our experiments.

Although simulated and with highly regular behaviors, the nice thing about the bAbI dialogue dataset is that it comes with an out-of-vocabulary (OOV) test set in which the entities (locations and food types) are not seen in any training dialogues. This poses exactly the same problem we are trying to address in this paper, namely predicting the API call parameters when they are not only unseen but also unknown to the system. Many of the previous E2E approaches simplifies the prediction problem as a selection among all API calls appeared in the entire dataset, thus bypassing the problem of tracking unknown dialogue states explicitly, although we believe it is not a realistic simplification.

\subsection{Model and Training Details}
\label{subsec:details}
The proposed model is implemented in TensorFlow. We use the provided development set to tune the hyper-parameters, track the training progress and select the best performing model for reporting the accuracy on test sets. The joint architecture is trained separately for each slot type by minimizing the sum of the cross entropy loss from the PtrNet and the classifier. Mini-batch SGD with a batch size of 50 and Adam optimizer~\citep{Adam:14} is used for training.

Each word is mapped to a {\it randomly initialized} 100 dimensional embedding and each dialogue instance is represented as a 540 * 100 dimensional vector with zero paddings on the left when necessary. Instead of the using the raw word sequences, the system utterances are replaced by the more succinct and consistent dialogue act representations such as ``request slot food". One layer of LSTM is used with a state size of 200 (additional layers did not help noticeably). Standard dropout with a keep probability of 0.5 is performed for training at the input and output of the LSTM cells. To keep it simple, targeted dropout is done only once for the entire training set before training begins, the dataset is therefore static across epochs.

To train the PtrNet, the location of the reference slot value in the dialogue needs to be provided. It does not require manual labeling though, and we simply use the {\it last} occurrence of the reference slot value in the dialogue history as the reference location. The occurrence is found via exact string match and the two most frequent spelling variations, ``moderate'' and ``moderately", ``center" and ``centre" are considered equivalent. If no occurrence exists in a training instance (due to ASR errors or rephrasing), it will not be used for training the PtrNet. 

On the other hand, the classifier serves as a gatekeeper that decides which slot values should be handed over to the PtrNet. On the bAbI dataset, there are zero non-pointable slots, and therefore everything is handled by the PtrNet. On DSTC2, we train the classifier to perform a {\it three-way} classification that determines if the slot values is none, dontcare or other. As we have described, other slot values can also become non-pointable in the actual dialogue: Those resulting from different surface forms are usually easier to handle, all we need is an extra post-processing step to normalize the value; The ones caused by ASR errors though, are much more challenging. One can argue that a classifier may be better equipped for these cases since it does not require locating the actual values in the word sequence, but unless there are consistent misrecognition patterns, they are difficult to handle for either the classifier or the PtrNet.

The non-pointable values in DSTC2, besides none and dontcare, are predominantly due to recognition errors, and we decide not to do anything specific about them -- the PtrNet is tasked with processing these misrecognized utterances, and no normalization (except for ``moderately" and ``center") is performed on the network output for computing the accuracy. \footnote{Non-pointable values besides none and dontcare constitute 9.7\% of {\it food}, 7.6\% of {\it location} and 4.7\% of {\it price} on the test data, effectively setting an upper bound on the accuracy.}

\subsection{Evaluation Setup}
The DSTC2 dataset is a standard benchmark for the task, we therefore compare the joint goal accuracy (a turn is considered correct if values are predicted correctly for all slots) of the proposed model with previous reported numbers to show the efficacy of our approach under regular circumstances, i.e. all slot values are known and observed in training. However, it is not our goal to outperform all previous DST systems -- the main theme is that our technique allows identifying unknown slot values effectively and even if used in the standard setting, our model yields state-of-the-art results. 

Measuring the accuracy on unknown slot values, however, does not have well-established baselines in the literature. Most previous systems are not concerned with this problem, and many of them are inherently not capable of outputting unknown values. So instead of comparisons with previous techniques, we will focus on demonstrating how this could be a serious problem tracking unknown slot values and how the targeted dropout can improve things drastically.

\subsection{Results}
The joint goal accuracy on the standard DSTC2 test set is shown in Table~\ref{tab:dstc_acc} comparing our PtrNet based model against various previous reported baselines.

\begin{table}[t!]
\begin{center}
\begin{tabular}{|c|c|}
\hline \bf Models & \bf Joint Acc.  \\ 
\hline
Delexicalizaed RNN  & 69.1 \\
Delexicalizaed RNN + semdict & 72.9 \\
NBT-DNN & 72.6 \\
NBT-CNN & 73.4 \\
MemN2N & 74.0 \\
Scalable Multi-domain DST & 70.3\\
\bf PtrNet & \bf 72.1 \\
\hline
\end{tabular}
\end{center}
\caption{Joint goal accuracy on DSTC2 test set vs. various approaches as reported in the literature. }
\label{tab:dstc_acc}
\end{table}

It is important to emphasize that the PtrNet model is an E2E model without using any SLU output and makes use of only the 1-best ASR hypothesis without any confidence measure for testing. Although more sophisticated DST models sometimes demonstrate better accuracy, our PtrNet model holds various advantages against all baseline models: In comparison with our approach, the delexicalized RNN models~\citep{Henderson:14b, Henderson:14c} utilize the n-best list and/or the SLU output; The NBT~\citep{Mrksic:17} and MemN2N~\citep{Perez:17} models are E2E but both depend on candidate lists as given and hence are not designed to handle unknown (different from unseen) slot values; The scalable DST model~\citep{Rastogi:18}, although addressing the same problem of unbounded value set, relies on SLU to generate value candidates, and also does not perform equally well on the standard test set.  

\begin{figure}
\centering
        \includegraphics[totalheight=5cm]{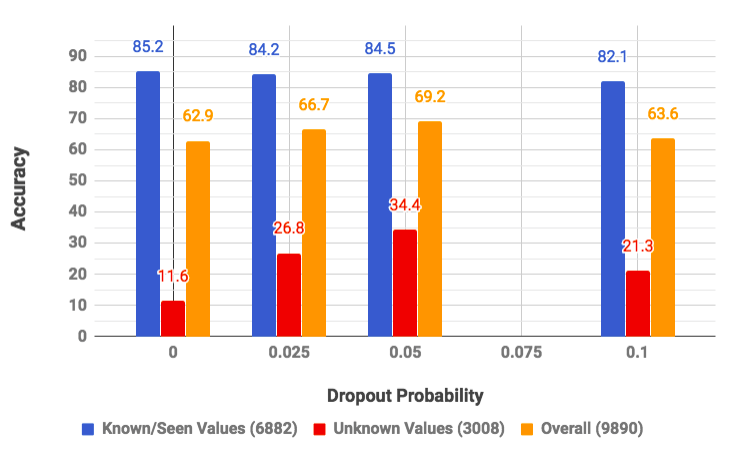}
    \caption{Accuracy of known/seen and unknown food types on the modified DSTC2 dataset with different dropout probabilities.}
    \label{fig:unk_drop}
\end{figure}

On the modified DSTC2 dataset with the reduced training set, the accuracy of the known/seen and unknown food types is shown in Figure~\ref{fig:unk_drop}. The standard training process with no targeted dropout performs poorly when the food types are not known beforehand, epitomizing the often overlooked challenge of handling unknown slot values. With a small dropout probability of 5\%, the accuracy on unknown values essentially increases by three times (from 11.6\% to 34.4\%), while the accuracy on other values remains roughly the same.

Similar observations can also be made on the bAbI dataset predicting OOV API parameters (Table~\ref{tab:babi}). While the dataset is quite artificial and in most cases we can achieve perfect accuracy on the regular test set, the OOV parameter values are not nearly as easy to predict. The targeted dropout however, allows us to bridge the accuracy gap entirely.

\begin{table}[t!]
\begin{center}
\begin{tabular}{|c|c|c|c|c|}
\hline
& \multicolumn{2}{|c|}{\bf Regular Test } &  \multicolumn{2}{|c|}{\bf OOV Test }\\ 
\cline{2-5}
& p=0 & p=0.1 & p=0 & p=0.1\\
\hline
food & 100 & 100 & 86.2 & 100\\
\hline
location & 100 & 100 & 74.7 & 99.6\\
\hline
\end{tabular}
\end{center}
\caption{Accuracy of predicting regular and OOV {\it food} and {\it location} parameters in bAbI (Task 5) API calls w/ (p=0.1) and w/o (p=0) targeted dropout. }
\label{tab:babi}
\end{table}

\section{Conclusion}
\label{sec:conclusion}
An E2E dialogue state tracker is introduced based on the pointer network. The model outputs slot values in an extractive fashion similar to the slot filling task in SLU. We also add a jointly trained classification component to combine with the pointer network, forming a hybrid architecture that not only achieves state-of-the-art accuracy on the DSTC2 dataset, but also more importantly is able to handle unknown slot values, which is a problem often neglected although particularly valuable in real world situations. A feature dropout trick is also described and proves to be particularly effective.

\section*{Acknowledgments}
We are grateful to the anonymous reviewers for their insightful comments. We also would like to thank Mei-Yuh Hwang for helpful discussions.
\bibliography{acl2018}
\bibliographystyle{acl_natbib}

\appendix
\end{document}